# Evaluation of Denoising Techniques for EOG signals based on SNR Estimation

Anirban Dasgupta[1], Suvodip Chakrborty[2], Aritra Chaudhuri[3], Aurobinda Routray[4]
IIT Kharagpur, India
[1]anirban1828@gmail.com, [2]suvodip.107019@gmail.com, [3]aritra00chaudhuri@gmail.com, [4]aroutray@ee.iitkgp.ernet.in

*Abstract*— **This paper evaluates four algorithms for denoising raw Electrooculography (EOG) data based on the Signal to Noise Ratio (SNR). The SNR is computed using the eigenvalue method. The filtering algorithms are a) Finite Impulse Response (FIR) bandpass filters, b) Stationary Wavelet Transform, c) Empirical Mode Decomposition (EMD) d) FIR Median Hybrid Filters. An EOG dataset has been prepared where the subject is asked to perform letter cancelation test on 20 subjects.**

## I. Introduction

Electrooculography (EOG) is a bio-signal that refers to the corneo-retinal standing potential[1]. This potential occurs whenever there is an eye movement. Usually, the retina has a negative bio-electric potential w.r.t the cornea [2]. The voltage for the horizontal eye movement is up to 16μV whereas it is 14μV for the vertical movement of the eye per 1°. The amplitude of EOG ranges between 50 to 3500 μV, and its frequency components go from 0 to 100Hz [1].

EOG signals are usually contaminated with noise, and hence it introduces errors in parameter estimation[3]. The major interferences in the capture of EOG are:

- Sensor noise
- Power-line noise
- Electroencephalography
- Electromyography
- Electrical network
- Speech
- Blink

Efficient removal of such noise is a challenging issue to the biomedical signal processing research community. The earlier works that aim at processing EOG signals mainly rely on low pass and notch filtering[1], [4], [5]. There has been substantial research on filtering EEG signals based on Empirical Mode Decomposition[6], Wavelet based Denoising[7],[8] and Kalman Filtering[9]. However, EOG has not obtained much attention using these methods.

This work aims at evaluating some of the popular state-of-the-art techniques for denoising EOG. The performance of the denoising technique has been evaluated using the SNR estimation.

The paper organization is as follows. Section II describes the capture of EOG signals. Section III aims at the denoising techniques. Section IV deals with SNR estimation process. Section V provides the results. Section VI concludes the paper.

## II. Data Capture

We conducted an experiment with 20 subjects to create the EOG database. The EOG was recorded using a polysomnography machine.

| S | G | H | K | B | T | R | H | I | O | V | P | Y |
|---|---|---|---|---|---|---|---|---|---|---|---|---|
| S | T | H | O | P | N | H | Y | U | I | T | R | B |
| T | L | I | K | B | G | T | R | P | D | H | G | P |
| S | F | M | H | Y | U | H | G | H | J | D | S | R |
| T | S | C | V | H | A | M | S | A | R | H | D | S |
| P | L | F | V | U | I | S | B | V | C | X | Z | R |
| S | D | T | G | E | O | H | J | B | U | Y | R | S |
| R | H | B | H | D | S | P | C | T | R | S | J | Z |
| X | C | V | B | S | H | A | O | A | H | T | R | D |
| S | P | S | G | S | H | L | F | U | V | P | D | P |
| O | G | V | Y | R | T | F | I | N | H | B | H | T |
| H | H | U | I | B | G | H | T | R | D | P | S | C |
| S | O | N | H | Y | T | G | H | R | S | D | V | T |

Fig. 1 Sample Set for Letter Counting Test

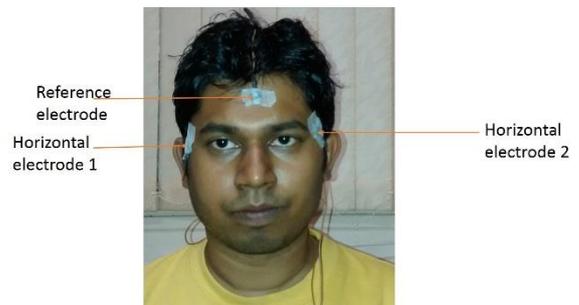

Fig. 2 Placement of EOG electrodes

The subjects were given video as well as verbal instructions regarding the tasks they had to perform. Informed consent and ethical approval were obtained from the concerned authorities respectively before the experiment.

The bipolar EOG electrodes [3] were placed near the canthus of both the eyes, with the reference electrode being placed in the middle of the forehead. Fig. 2 shows the placement of electrodes. A 13×13 array of English alphabets was placed in front of the subject, and the subject is asked to find out the number of occurrences of the alphabet 'A' from the list. The list is shown in Fig. 1. The EOG data was sampled at 256 Hz.

## III. DENOISING TECHNIQUES

The EOG time series data being non-stationary, has to be unbiased for the pre-processing[10]. We have employed the mean removal technique for the purpose. For the EOG sequence $x(n)$ for a particular window size $N$ and its mean $\mu(n)$, the stationary approximation is obtained as

$$x_s(n) = x(n) - \mu(n) \qquad (1)$$

Here we have selected a moving window size of 256 samples empirically and with 25% overlap. The EOG signal $x_s(n)$ is considered as it has been corrupted by additive white noise during the process of signal acquisition. The corrupted EOG signal (observed) is given as

$$x_s(n) = y(n) + \eta(n) \qquad (2)$$

Where $y(n)$ is uncontaminated EOG signal, $\eta(n)$ represents the statistically independently and identically distributed (i.i.d.) white Gaussian noise with $N(0, \sigma^2)$. Here the problem is to remove or attenuate $\eta(n)$ to the maximum extent from the output signal $x_s(n)$. In this work, we have employed three different techniques for noise removal as follows:

*a) Band Pass FIR Filtering*

The band-pass FIR filter is designed using a rectangular window. The filter specifications are as follows:
- Order: 10
- Normalized Low frequency: 0.02
- Normalized High frequency: 0.5
- Method: Least Squares

$$\hat{y}_1(n) = x_s(n) * h(n) \qquad (3)$$

$\hat{y}_1(n)$ gives an estimate of the filtered sequence with $h(n)$ as the filter coefficients and $x_s(n)$ as the input sequence, with $*$ denoting convolution sum.

*b) Empirical Mode Decomposition (EMD)*

EMD is a data-driven filtering technique that decomposes a time series into various components called Intrinsic Mode Functions (IMF)[11]. These components form a complete and nearly orthogonal basis for the original time-series. The algorithm is listed in Table 1. In the present work, we have obtained 11 IMFs from the EOG data, and finally selecting the IMFs 2 to 9 empirically, we obtain the filtered sequence $\hat{y}_2(n)$.

Table 1: EMD for noise removal from EOG data

| Algorithm: EMD |
|---|
| **Input:** Time series |
| **Output:** IMFs |
| 1. Identify the extrema (maxima and minima) of the signal $x_s(n)$. |
| 2. Interpolate the maxima of the $x_s(n)$ by using a natural cubic spine to obtain $e_{max}(n)$ |
| 3. Similarly, interpolate the minima of the $x_s(n)$ by using a natural cubic spine to obtain $e_{min}(n)$ |
| 4. The mean curve is then obtained as $m(n) = \frac{e_{min}(n) + e_{max}(n)}{2}$ |
| 5. The detail is obtained as $d(n) = x_s(n) - m(n)$ |
| 6. The residual $m(n)$ is iterated |
| 7. The filtered sequence $\hat{y}_2(n)$ is obtained by summing up the necessary IMFs |

*c) Stationary Wavelet Transform (SWT) based Filtering*

SWT is a type of wavelet transform used for denoising biomedical signals[12]. The Discrete Wavelet Transform (DWT) does not preserve translation invariance due to sub-sampling operations in the pyramidal algorithm. The SWT preserves the property that a translation of the original signal does not necessarily imply a translation of the corresponding wavelet coefficients and hence has been employed in this work.

Let $\psi(n)$ be the mother wavelet with scales $a$ and positions $b$.

$$\psi(n) = 2^{a/2}\psi(2^a n - b) \qquad (4)$$

Here $a$ and $b$ takes on powers of 2. The high frequency and low frequency components of the EOG signal $x_s(n)$ match the contracted and dilated versions of the wavelet function, respectively. The denoising algorithm is provided in Table 2.

Table 2: SWT for noise removal from EOG data

| Algorithm: SWT |
|---|
| **Input:** Time series |
| **Output:** SWT component |
| 1. Application of SWT to the contaminated EOG with a specific wavelet as basis function and decomposition up to 6 levels. |
| 2. Application of Soft or Hard Threshold. |
| 3. Reconstruction of decomposed signal to obtain denoised EOG signal, $\hat{y}_3(n)$. |

The SWT yielded $\hat{y}_3(n)$, as the filtered reconstruction.

$$\hat{y}_3(n) = \sum_{m \in Z} <x, \psi_m> \cdot \psi_m(t) \qquad (5)$$

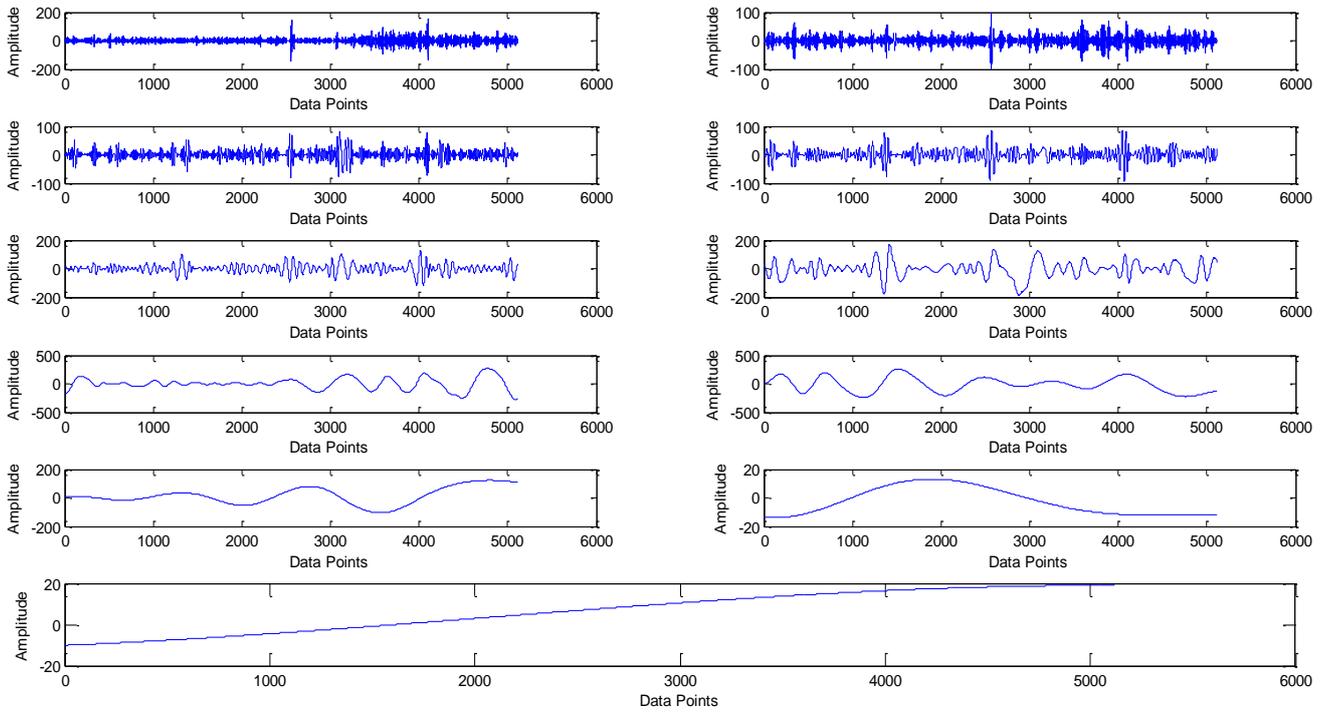

Fig. 3 The IMFs found from the EMD

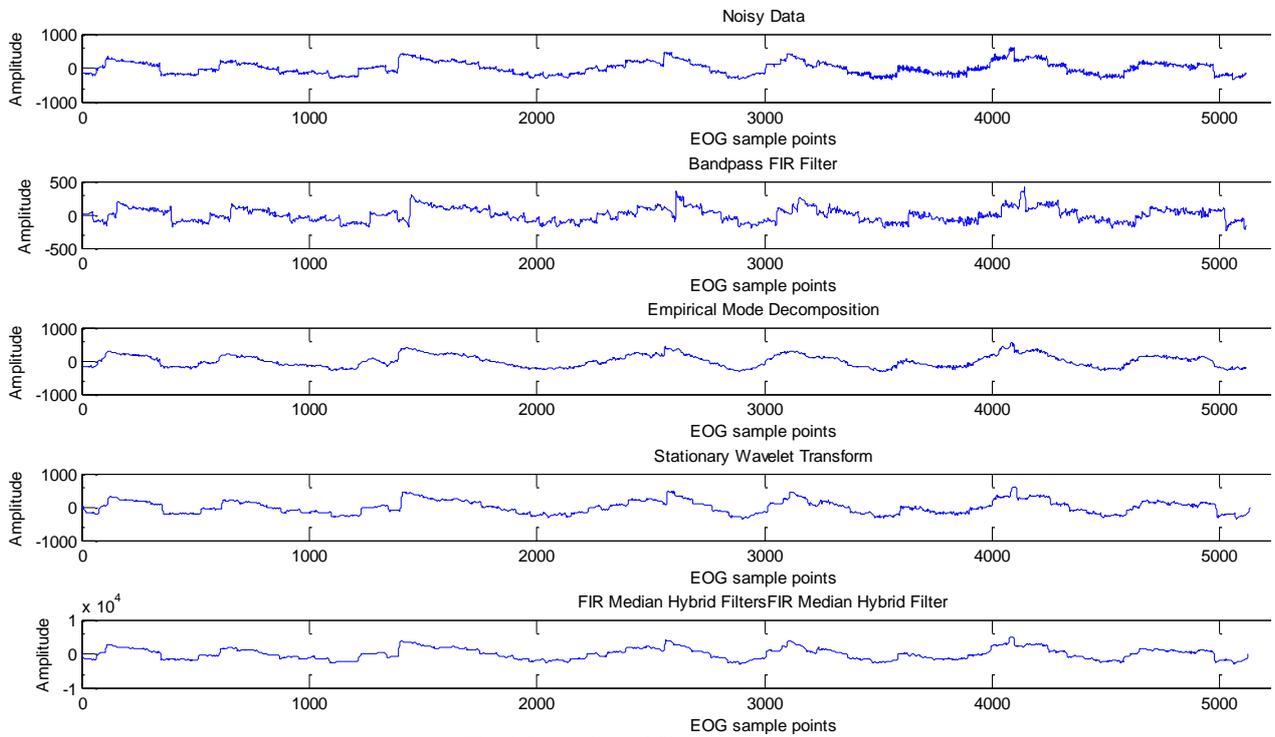

Fig. 4 Comparison of filtering methods

*d)*      *FIR Median Hybrid Filters*

These filters combine the properties of the FIR filters [2] for noise removal and the capability of median filters of preserving edges. Such a filter consists of $M$ subfilters.

$$\hat{y}_4(n) = median\{w_1p_1, w_2p_2, w_3p_3, w_4p_4, w_5p_5\} \quad (6)$$

In the above equation, $median\{\}$ represents the sample median of the sequence and $w_i$'s are the corresponding

weights. Here the $p_i$'s are functions of $n$, and are obtained using the following equations:

$$p_1 = \frac{1}{L}\sum_{i=0}^{L-1} x_s(n) \quad (7)$$

$$p_2 = \sum_{i=0}^{L-1} h(i)x_s(n-i) \quad (8)$$

$$p_3 = x_s(n) \quad (9)$$

$$p_4 = \sum_{i=0}^{L-1} h(i)x_s(n) \quad (10)$$

$$p_5 = \frac{1}{L}\sum_{i=0}^{L-1} x_s(n-i) \quad (11)$$

$\hat{y}_4(n)$ is the output sequence using the FIR median hybrid filters.

## IV. SNR ESTIMATION

An estimation of SNR, $S$ is required to evaluate the performance of the algorithms [12]. First, the noise from each algorithm is obtained as

$$\eta_i = y(n) - \hat{y}_i(n) \quad (12)$$

The noise covariance of the noise subspace by

$$cov(\eta) = \eta'\eta \quad (13)$$

Here $\eta$ is the vector $\eta = [\eta(0)\eta(1)\eta(2)\dots\eta(n)]'$. The Eigen values of the noise covariance matrix is calculated, Next the signal covariance matrix is calculated using,

$$cov(y(n)) = y'y \quad (14)$$

The Eigenvalues of the signal covariance matrix are calculated. Now we calculate $S$ by,

$$S = log(\frac{max(eig(signal_{cov})) - max(eig(noise_{cov}))}{max(eig(noise_{cov}))}) \quad (15)$$

## V. RESULTS

The algorithm was tested on the captured data. The results are tabulated in Table 3.

Table 3: Comparison of the denoising algorithms

| Method | SNR (in dB) | Time (in ms) |
|---|---|---|
| Band Pass FIR | 21.75 | 22.4 |
| EMD | 31.12 | 33.7 |
| SWT | 24.23 | 31.76 |
| FIR Median Hybrid Filter | 25.15 | 29.76 |

## VI. CONCLUSION

Denoising of EOG signals is a considerably challenging problem of biomedical signal processing, as it is essentially a mixture of an unknown amount of correlated noise with a non-stationary signal. Four algorithms for denoising EOG signal were evaluated using the SNR as a performance index. Our main observation supports that EMD provides the best SNR value as compared to the other techniques tested. Bandpass FIR filter shows considerable accuracy with possibly the best processing speed out of all filtering techniques.

## VII. ACKNOWLEDGEMENT

The funds received from the Ministry of Human Resource Development, Government of India are thankfully acknowledged